\documentclass[runningheads]{llncs}
\usepackage{graphicx}

\usepackage[dvipsnames]{xcolor}
\usepackage{tikz}
\usepackage{comment}
\usepackage{amsmath,amssymb} 
\usepackage{color}

\usepackage[accsupp]{axessibility}  

\usepackage[pagebackref=true,breaklinks=true,colorlinks,bookmarks=false]{hyperref}

\usepackage{subcaption, booktabs, multirow}
\usepackage{definitions}

\begin{document}
\pagestyle{headings}
\mainmatter
\def\ECCVSubNumber{199}  

\title{Hydra Attention:\\Efficient Attention with Many Heads}


\titlerunning{Hydra Attention}
%
\author{Daniel Bolya\inst{1,2}\thanks{{This work was done under an internship at Meta AI.}} \and
Cheng-Yang Fu\inst{2} \and
Xiaoliang Dai\inst{2} \and
Peizhao Zhang\inst{2} \and
Judy Hoffman\inst{1}}
\authorrunning{D. Bolya et al.}
%
\institute{Georgia Tech\\
\email{\{dbolya,judy\}@gatech.edu} \and
Meta AI\\
\email{\{chengyangfu,xiaoliangdai,stzpz\}@fb.com}}
\maketitle

\begin{abstract}
While transformers have begun to dominate many tasks in vision, applying them to large images is still computationally difficult. A large reason for this is that self-attention scales quadratically with the number of tokens, which in turn, scales quadratically with the image size. On larger images (e.g., 1080p), over 60\% of the total computation in the network is spent solely on creating and applying attention matrices. We take a step toward solving this issue by introducing Hydra Attention, an extremely efficient attention operation for Vision Transformers (ViTs). Paradoxically, this efficiency comes from taking multi-head attention to its extreme: by using as many attention heads as there are features, Hydra Attention is computationally linear in both tokens and features with no hidden constants, making it significantly faster than standard self-attention in an off-the-shelf ViT-B/16 by a factor of the token count. Moreover, Hydra Attention retains high accuracy on ImageNet and, in some cases, actually \textit{improves} it.

\end{abstract}

\begin{figure}[t]
\begin{center}
	\includegraphics[width=1\linewidth]{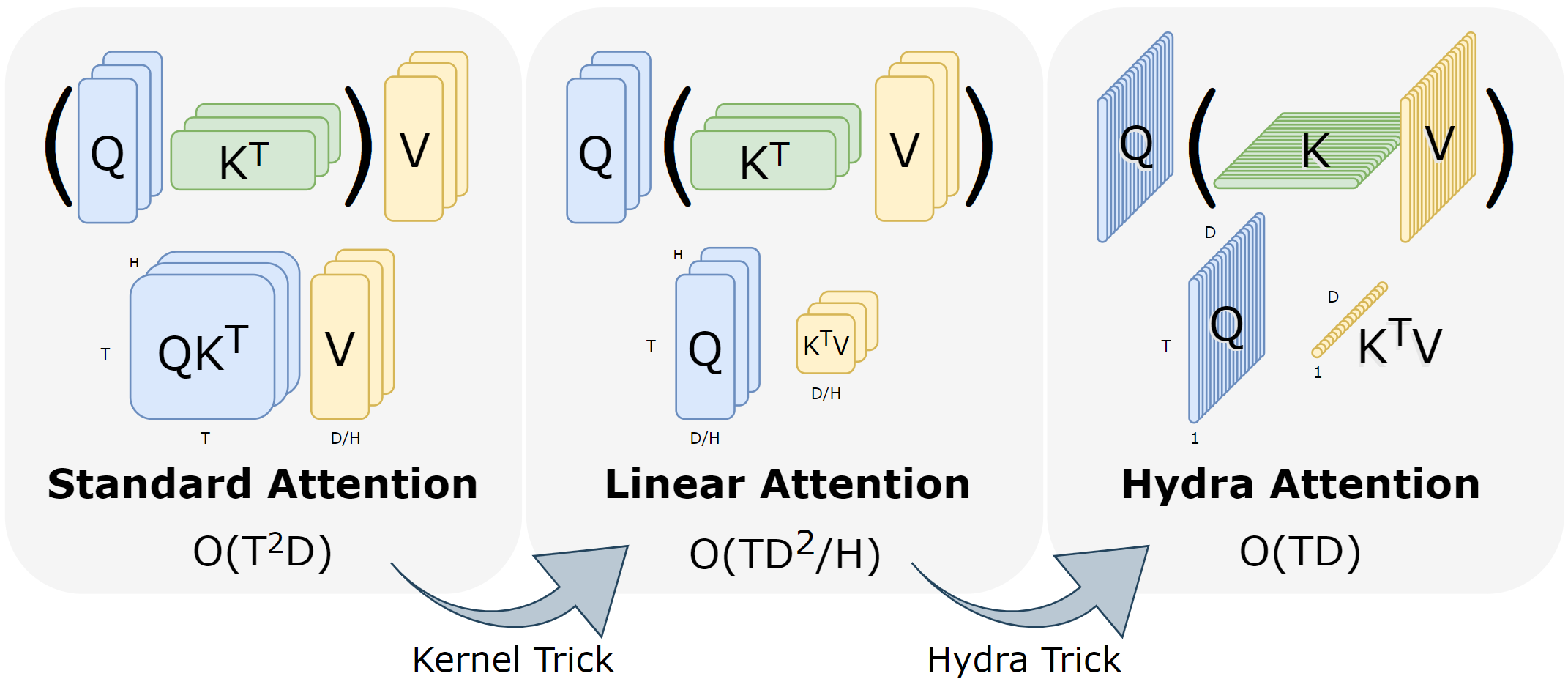}
    \caption{\textbf{Hydra Attention.} Standard attention \cite{attnisallyouneed} scales with the square of the number of tokens $T$. Using a decomposable kernel, we can rearrange the order of operations as in \cite{linearattn_rnn} such that attention scales with the square of features $D$ instead. Our Hydra attention goes one step further by maximizing the number of attention heads $H$, resulting in an $O(TD)$ operation in both space and time.} 
	\label{fig:method}
\end{center}

\end{figure}
\section{Introduction}

Because of their generality and high capacity to learn from large amounts of data, transformers \cite{attnisallyouneed} have been a dominant force in natural language processing (NLP) for the last couple of years \cite{bert,gpt2,palm}. And now, with the introduction of Vision Transformers (ViTs) \cite{vit}, the same takeover is happening in vision.

Yet, unlike in NLP, the pure instantiation of transformers that can be seen in NLP with BERT \cite{bert} or in vision with ViT \cite{vit} are not the force dominating computer vision tasks. Instead, much more vision-specialized attention-based architectures such as Swin \cite{swin} or MViT \cite{mvit,mvit2} or attention-conv mixtures like LeViT \cite{levit} are being used instead.

The primary reason behind this discrepancy is efficiency: specialized vision transformers can perform better with less compute---either by adding conv layers, by using vision-specific local window attention, or by using some other way to cheaply add visual inductive bias. While pure ViTs can perform well at scale (90.45\% top-1 on ImageNet \cite{scalingvits}), the primary mechanism of a pure transformer---multihead self-attention \cite{attnisallyouneed}---can be an extreme bottleneck when applying a model on the large images required by several downstream tasks.

In fact, when applying an off-the-shelf ViT on 1080p images, common for benchmark tasks such as segmentation (e.g., CityScapes \cite{cityscapes}), 60\% of the total computation in the network (see Tab.~\ref{tab:flops}) is spent simply on creating and applying attention matrices for self-attention, compared to 4\% on $224 \times 224$ ImageNet \cite{imagenet} images. In a pure transformer, these attention matrices scale computationally with the square of tokens, which can already be prohibitively expensive (such as with long sentences in NLP). But in a ViT, the problem is compounded further by the tokens scaling with the square of the image size, meaning doubling the image size increases the computation in attention by a factor of 16.

There are already a wealth of techniques that have been explored to address this problem in the NLP space. Several works have introduced ``linear'' attention (in terms of tokens) either by re-arranging the order of computation using a ``kernel trick'' \cite{performers,linearattn_efficient,linearattn_rnn,randomfeatureattn} or projecting to a token-independent low-rank space \cite{linformer,performers,randomfeatureattn}, some doing both. However, most of these ``linear'' attention methods trade computation across the tokens for computation across the features, making them rather expensive. In fact, recently, Flash Attention \cite{flashattn} has shown that an IO efficient implementation of multihead self-attention can outperform most of these ``linear'' attention methods even with token counts in the thousands.

A few works have attempted efficient attention in the vision space, too, but none have been explored on their own in a traditional ViT shell. PolyNL \cite{polynl} treats attention as an efficient third-order polynomial, but this hasn't yet been explored in a ViT architecture. Attention Free Transformer \cite{aft} has an AFT-Simple variant that is similarly efficient, but it performs poorly in a pure ViT and requires extra support from convs and position encodings. We test both of these methods in a standard DeiT \cite{deit} shell (see Tab.~\ref{tab:kernel_choice}), and find that both methods, while efficient, result in a significant accuracy drop. Thus, there is room in the literature for a truly efficient, accurate, and general replacement for multihead self-attention.

To that extent, we introduce Hydra Attention (see Fig.~\ref{fig:method}). Our method results from a somewhat paradoxical behavior in linear attention: with standard multihead self-attention, adding more heads to the model keeps the amount of computation the same. However, after changing the order of operations in linear attention, adding more heads actually \textit{reduces} the compute cost of the layer. We take this observation to its extreme by setting the number of heads in the model to be equal to the number of features, thereby creating an attention module that's computationally linear with respect to both tokens and features.

Not only is Hydra Attention a more general formulation of previous efficient attention works (see Sec.~\ref{subsec:other_works}), but when using the right kernel, it can be significantly more accurate (see Tab.~\ref{tab:kernel_choice}). In fact, when mixed with standard multi-head attention, Hydra Attention can actually \textit{increase} the accuracy of a baseline DeiT-B model while being faster (see Fig.~\ref{fig:backward_forward_scan}). And by being derived from multihead attention, our method retains several of attention's nice properties, such as explainability (see Fig.~\ref{fig:hydra_attn_vis}) and generality to different tasks.

However, while Hydra Attention is general and efficient for large images, in this paper we focus solely on ImageNet \cite{imagenet} classification using DeiT-B \cite{deit}, which traditionally uses smaller $224 \times 224$ and $384 \times 384$ images. While the efficiency gains aren't as much here (10-27\% based on image size), other efficient attention methods (e.g., \cite{aft,polynl}) already suffer from huge accuracy drops in this regime (see Tab.~\ref{tab:kernel_choice}), whereas Hydra Attention does not. We hope Hydra Attention can become a stepping stone for general, pure transformers with large numbers of tokens in the future.

Our contributions are as follows: we perform a study to validate how many heads a transformer can have (Fig.~\ref{fig:head_experiment}) and find that 12 is the limit for softmax attention, but with the right kernel, any number is feasible. Then we use that observation to introduce Hydra Attention (Sec.~\ref{sec:method}) for pure transformers by increasing the number of heads in multihead self-attention. We then analyze the action of Hydra Attention mathematically (Sec.~\ref{subsec:hydra_trick}) and introduce a method to visualize its focus (Fig.~\ref{fig:hydra_attn_vis}). Finally, we find that by replacing specific attention layers with Hydra Attention (Fig.~\ref{fig:backward_forward_scan}), we can either \textit{improve} accuracy by 1\% or match the accuracy of the baseline, while producing a strictly faster model using DeiT-B \cite{deit} on ImageNet-1k \cite{imagenet}.

\section{Related Work} \label{sec:related_work}
In this paper, our goal is to speed up the inference time of a transformer by removing the token squared computation bottleneck in multihead self-attention.

\paragraph{Efficient Attention}
Multihead Self-Attention \cite{attnisallyouneed} is a notoriously slow operation, and there have been plenty of works trying to address its computational shortcomings in different domains.

In NLP, several works approximate attention with a decomposable kernel function \cite{performers,linearattn_efficient,linearattn_rnn,randomfeatureattn}. This ``kernel trick'' allows them to reorder the matrix multiplications to be quadratic in terms of features instead of tokens. Some of these methods go further and reduce the dimensionality of this matrix multiplication through a projection to a low rank space \cite{linformer,performers,randomfeatureattn}. However, these ``linear'' attention methods trade computation across the tokens for computation across the features, which can make them expensive. In fact, in the domain of this paper (ImageNet classification), there aren't enough tokens to justify these approaches and most of them produce a \textit{slower} model. And even with thousands of tokens, Flash Attention \cite{flashattn} has shown that an IO-aware implementation of multihead self-attention can actually outspeed even the fastest of these methods.

But reordering operations isn't the only way to speed up attention. In fact, the most common way to ``linearize'' attention in vision is by using local window attention (e.g., \cite{swin,vit-p,vitdet}). This is indeed computationally linear with respect to the number of tokens, but local window attention can be difficult to compute (especially in the case of Swin \cite{swin}) and this is only possible with dense, spatially ordered modalities such as images and videos.

Our goal is instead to produce a linear attention method that is efficient, fast to compute, and general across several different modalities.

\paragraph{Efficient Transformers}
Replacing the attention module is not the only way to speed up the inference time of a transformer. In fact, depending on the task and the number of tokens, other efficient transformer methods can be more desirable. For instance, attention only accounts for 4\% of the total network computation on ImageNet \cite{imagenet} classification, meaning 4\% is the maximum obtainable speed-up if only attention is modified.

There are several efficient vision transformers that mix convs and attention together to create a more efficient end product, such as LeViT \cite{levit}, MobileViT \cite{mobilevit}, Mobile-Former \cite{mobileformer}, and LVT \cite{lvt}. All of these are a valid strategy for images, and we view them as adjacent techniques. Other vision-specific attention papers such as \cite{aft,polynl} use convolutions in addition to their efficient attention, making it difficult to discern whether the improvement comes from the attention method or the introduction of convolution.

In this paper, we make no modifications to the underlying ViT architecture except to swap multihead self-attention for Hydra Attention in order to clearly isolate its impact on performance.

\paragraph{Multihead Attention}
Hydra Attention relies on increasing the number of heads used in multihead attention. Interestingly enough, since its introduction in \cite{attnisallyouneed}, the number of heads used for multihead attention has not been explored in much depth. Some studies have been done on pruning attention heads \cite{analyzingmultiheadattn,sixteenheadsbetterthanone}, however all studies have been in the direction in reducing the number of heads. In fact, even with ViT-G, the largest ViT models explored in \cite{scalingvits}, the authors only use 16 attention heads. Thus, we conduct this study ourselves in Fig.~\ref{fig:head_experiment}.

\section{Hydra Attention} \label{sec:method}
Standard multihead self-attention \cite{attnisallyouneed} scales quadratically with the number of tokens in an image. More concretely, if $T$ is the number of tokens and $D$ is the number of feature dimensions, then creating and applying an attention matrix are both $O(T^2 D)$. This poses a problem, then, when $T$ is large (as it is the case with large images), as this operation can quickly become computationally infeasible.

\subsection{The Kernel Trick}
As discussed in Sec.~\ref{sec:related_work}, many works \cite{performers,linearattn_efficient,linearattn_rnn,randomfeatureattn} have already attempted to address this by introducing ``linear'' attention. Given queries $Q$, keys $K$, and values $V$ in $\mathbb{R}^{T \times D}$, standard softmax self-attention is computed as
\begin{equation}
    A(Q, K, V) = \text{softmax}\left(\frac{QK^T}{\sqrt{D}}\right) V \label{eq:sigmoid_attn}
\end{equation}
Computing $QK^T$ is $O(T^2 D)$ and creates a $T\times T$ matrix, which scales poorly with $T$. As in \cite{linearattn_rnn}, we can generalize this operation by treating $\text{softmax}(\cdot)$ as a pairwise similarity between $Q$ and $K$. That is, for some similarity function $\text{sim}(\cdot)$, we can write
\begin{equation}
    A(Q, K, V) = \text{sim}(Q, K) V
\end{equation}
If we then choose a decomposable kernel with feature representation $\phi(\cdot)$ such that $\text{sim}(x, y) = \phi(x)\phi(y)^T$, we can obtain
\begin{equation}
    A(Q, K, V; \phi) = \left(\phi(Q) \phi(K)^T\right) V
\end{equation}
Then by associativity, we can change the order of computation such that
\begin{equation}
    A(Q, K, V; \phi) = \phi(Q) \left(\phi(K)^T V\right) \label{eq:linear_attn}
\end{equation}
This allows us to compute $\phi(K)^T V$ first, leading to an operation that is $O(TD^2)$ and that creates a $D^2$ matrix instead of a $T^2$ one. Note this formulation differs slightly from \cite{linearattn_rnn}, in that we leave the normalization to the similarity function rather than make it explicit.

\subsection{Multi-Head Attention}
Despite being linear with respect to $T$, the result in Eq.~\ref{eq:linear_attn} is still undesirable: $D$ is typically large ($\geq768$) and so creating a $D \times D$ matrix and performing $O(TD^2)$ operations can still be quite costly. However, Eq.~\ref{eq:sigmoid_attn} through Eq.~\ref{eq:linear_attn} assume that we create one attention matrix, and thus have one ``head''.

In practice, most vision transformers use $H$ heads (typically between 6 and 16), where each head creates and applies its own attention matrix. Following \cite{attnisallyouneed}, each of heads operate on their own $D/H$ subset of features from $Q$, $K$, and $V$. Thus Eq.~\ref{eq:sigmoid_attn} becomes
\begin{equation}
    A(Q_h, K_h, V_h) = \text{softmax}\left(\frac{Q_hK_h^T}{\sqrt{D}}\right) V_h \qquad \forall h \in \{1, \ldots, H\} \label{eq:multihead_softmax_attn}
\end{equation}
where $Q_h, K_h, V_h \in \mathbb{R}^{T \times \frac{D}{H}}$. This keeps the total number of operations the same:
\begin{equation}
    O(H T^2 D/H) = O(T^2 D) \label{eq:msa_ops}
\end{equation}
The same is not true, however, for linear attention. Eq.~\ref{eq:linear_attn} becomes
\begin{equation}
    A(Q_h, K_h, V_h; \phi) = \phi(Q_h) \left(\phi(K_h)^T V_h\right) \qquad \forall h \in \{1, \ldots, H\} \label{eq:multihead_linear_attn}
\end{equation}
By computing attention in this way, adding heads actually \textit{decreases} the number of operations:
\begin{equation}
    O(H T (D/H)^2) = O(T D^2 / H) \label{eq:mla_ops}
\end{equation}

\begin{figure}[t]
\begin{center}
	\includegraphics[width=1\linewidth]{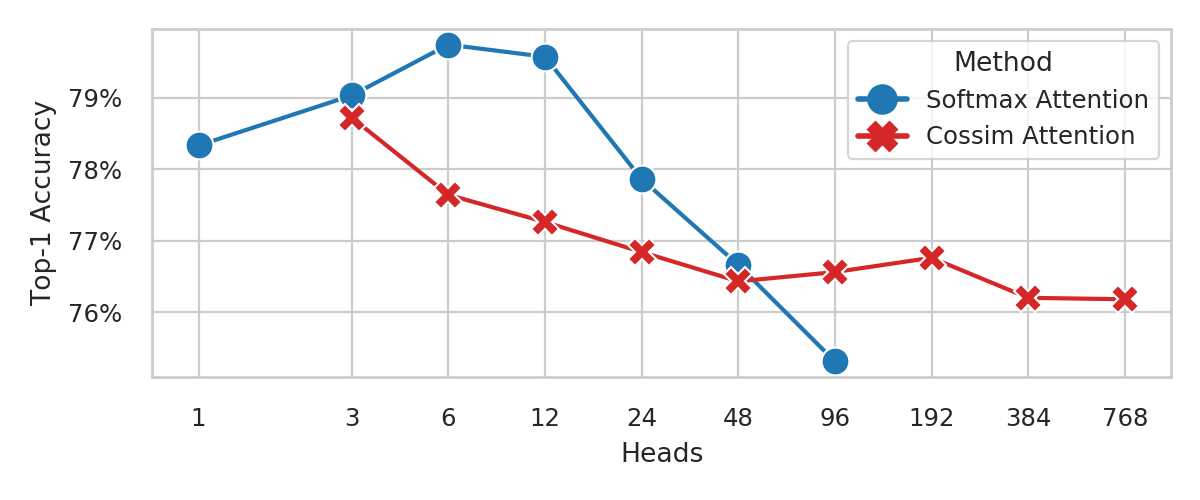}
    \caption{\textbf{Varying Heads.} We train a DeiT-B model on ImageNet-1k with different numbers of heads using either standard self-attention (blue) using softmax or multi-head linear attention (red) using cosine similarity. Results for standard self-attention ran out of memory for $H>96$ and multi-head linear attention for $H<3$. Softmax attention seems to crash in accuracy as we add more heads, while multi-head linear attention stays consistent. Note that $H$ must divide $D = 768$.}
	\label{fig:head_experiment}
\end{center}

\end{figure}

\subsection{Adding Heads}
Given Eq.~\ref{eq:mla_ops}, the more heads we add to the network, the faster multihead linear attention becomes. That begs the question, how many heads can we reasonably add, anyway? Most transformers in the wild use between 6 and 16 heads \cite{attnisallyouneed,bert,vit,scalingvits} depending on the number of features $D$, but what happens if you increase the number of heads beyond that?

To find out, we train DeiT-B \cite{deit} on ImageNet-1k \cite{imagenet} and vary the number of heads $H$ using either standard multi-head self-attention (Eq.~\ref{eq:multihead_softmax_attn}, MSA) with softmax or multi-head linear attention (Eq.~\ref{eq:multihead_linear_attn}, MLA) with cosine similarity, plotting the results in Fig.~\ref{fig:head_experiment}. In terms of memory usage, MSA runs out of memory when $H > 96$ and MLA runs out of memory when $H < 3$.

In terms of performance, while the accuracy for MSA tanks for $H > 12$, the accuracy for MLA with cosine similarity stays quite consistent all the way up to $H = 768$. Amazingly, at this number of heads, $H$ is equal to $D$, meaning each head has only a scalar features to work with!

\subsection{The Hydra Trick} \label{subsec:hydra_trick}
As shown in Fig.~\ref{fig:head_experiment}, it's feasible to scale $H$ up arbitrarily as long as the similarity function $\text{sim}(x, y)$ is not softmax. To exploit this, we introduce the ``hydra trick'', where we set $H = D$:
\begin{equation}
    A(Q_h, K_h, V_h; \phi) = \phi(Q_h) \left(\phi(K_h)^T V_h\right) \qquad \forall h \in \{1, \ldots, D\} \label{eq:multihead_linear_attn_hydra}
\end{equation}
In this case, each $Q_h, K_h, V_h$ is a column vector in $\mathbb{R}^{T \times 1}$. If we then vectorize the operation across the heads, we end up with
\begin{equation}
    \text{Hydra}(Q, K, V; \phi) = \phi(Q) \odot \sum_{t=1}^T \phi(K)^t \odot V^t \label{eq:hydra_attn}
\end{equation}
where $\odot$ denotes element-wise multiplication. Note there is a subtle difference between this vectorization and Eq.~\ref{eq:multihead_linear_attn_hydra}: $\phi$ is applied to the entirety of $Q$ and $K$, rather than to individual column vectors $Q_h$ and $K_h$. This is important because for each token, $Q_h$ and $K_h$ are scalars, and taking the similarity between two scalars is very restrictive (e.g., cosine similarity can only output -1, 0, or +1).

Also, while the derivation of Eq.~\ref{eq:hydra_attn} comes from multihead attention, it actually ends up performing something quite different: it first creates a global feature vector $\sum_{t=1}^T \phi(K)^t \odot V^t$ that aggregates information across all the tokens in the image. Then each $\phi(Q)$ gates the importance of this global feature for each output token. Thus, Hydra Attention mixes information through a global bottleneck, rather than doing explicit token-to-token mixing as in standard self-attention.

This results in a computational complexity of
\begin{equation}
    O(TD (D / H)) = O(TD) \label{eq:hydra_ops}
\end{equation}
leaving us with an efficient token mixing module that is linear with both the number of tokens and features in the model, and with no extra constants as in other linear attention methods (such as \cite{performers,linformer,linearattn_rnn}). Note that the space complexity of this technique is also $O(TD)$, which is important for real-world speed, where many operations are IO-bound (see \cite{flashattn}).

\subsection{Relation to Other Works} \label{subsec:other_works}
There are a few other $O(TD)$ attention candidates in the literature: Attention-Free Transformer \cite{aft} (specifically AFT-Simple) and PolyNL \cite{polynl}. In this section, we explore how Hydra Attention as described in Eq.~\ref{eq:hydra_attn} relates to each.

AFT-Simple \cite{aft} is described as
\begin{equation}
    \text{AFT-Simple}(Q, K, V) = \sigma(Q) \odot \sum_{t=1}^T{\text{softmax}(K)^t \odot V^t}
    \label{eq:aftsimple}
\end{equation}
where $\sigma(\cdot)$ denotes sigmoid. If we allow $\phi$ to vary between $Q$ and $K$, this is a direct specilization of Eq.~\ref{eq:hydra_attn} with $\phi(Q) = \sigma(Q)$ and $\phi(K) = \text{softmax}(K)$.

PolyNL \cite{polynl}, on the other hand, is described as
\begin{equation}
    \text{PolyNL}(X; W_1, W_2, W_3) = \left(X \odot \frac{1}{T} \sum_{t=1}^T{XW_1 \odot XW_2}\right) W_3
    \label{eq:polynl}
\end{equation}
If we denote $K = XW_1$ and $V = XW_2$, and let $\phi_\text{mean}(x) = x / \sqrt{T}$, we can write
\begin{equation}
    \text{PolyNL}(X; W_1, W_2, W_3) = \text{Hydra}(X, K, V; \phi_\text{mean}) W_3
\end{equation}
Thus, Hydra attention can be seen as a more general form of other $O(TD)$ attention methods.

\section{Experiments}
For all experiments, unless otherwise noted, we use DeiT-B \cite{deit} with default settings trained on ImageNet-1k \cite{imagenet} reported as Top-1 accuracy on the validation set.
When not specified, the function used for $\phi(\cdot)$ in Eq.~\ref{eq:hydra_attn} is L2 normalization such that $\text{sim}(\cdot, \cdot)$ is cosine similarity. To compute throughput, we sweep over several batch sizes and report the highest average throughput on 30 batches after 10 discarded warm-up iterations.



\begin{table}[t]
\begin{center}
	\begin{smalltable}{l c c c r}
	    \toprule
	    Method & Kernel & $\qquad\phi(Q)\qquad$ & $\qquad\phi(K)\qquad$ & Accuracy \\
	    \midrule
	    Hydra & Cosine Similarity & \multicolumn{2}{c}{$x / ||x||_2$} & {\bf 76.37} \\
	    Hydra & Mean & \multicolumn{2}{c}{$x / \sqrt{T}$} & 75.95 \\
	    Hydra & Tanh Softmax & $\text{tanh}(x)$ & $\text{softmax}(x)$ & 74.18 \\
	    \midrule
	    AFT-Simple \cite{aft} & Sigmoid Softmax & $\sigma(x)$ & $\text{softmax}(x)$ & 74.02 \\
	    PolyNL \cite{polynl} & Mean & \multicolumn{2}{c}{$x / \sqrt{T}$} & 73.96 \\
	    \bottomrule
	\end{smalltable}
    \caption{\textbf{Kernel Choice.} Here we vary the choice of kernel function through its feature representation $\phi(\cdot)$ in Eq.~\ref{eq:hydra_attn}. We also compare against AFT and PolyNL here as mentioned in Sec.~\ref{subsec:other_works}. Note that some kernels can be asymmetric, with different $\phi(Q)$ and $\phi(K)$. See the appendix for more kernels. }
	\label{tab:kernel_choice}
\end{center}

\end{table}
\subsection{The Choice of Kernel} \label{subsec:kernel_choice}
In most of our experiments, following \cite{linearattn_rnn} we use cosine similarity as our kernel function for Eq.~\ref{eq:hydra_attn}. In Tab.~\ref{tab:kernel_choice}, we explore other possible kernels, including those used by other candidate attention replacement methods as discussed in Sec.~\ref{subsec:other_works}. Yet, no kernel we test outperforms simple cosine similarity.

This might be because cosine similarity changes the nature of attention. With MSA (Eq.~\ref{eq:multihead_softmax_attn}), attention exclusively mixes information contained in $V$, as the mixing weights $\text{sim}(Q, K)$ must sum to 1. That's not the case when using cosine similarity or other unrestricted dot-product kernels like mean. And it turns out, these weights summing to 1 might not be a desirable property in the first place: AFT-Simple \cite{aft} as described in Eq.~\ref{eq:aftsimple} sets $\phi(Q) = \sigma(Q)$ and $\phi(K)=\text{softmax}(K)$, which is closer to a strict mixing of $V$, but the performance suffers as a result (see Tab.~\ref{tab:kernel_choice}).

We also test using $\text{tanh}(Q)$ instead of $\sigma(Q)$ to see if cosine similarity allowing the result to be negative was the reason, but that performs only slightly better than AFT-Simple. Thus, in this computationally constrained environment, it seems that leaving the kernel to be as unrestricted as possible while normalizing it in some way is important. We test several other kernels and note them in the appendix, but none outperform this simple technique.

\begin{figure}[t]
\begin{center}
	\includegraphics[width=1\linewidth]{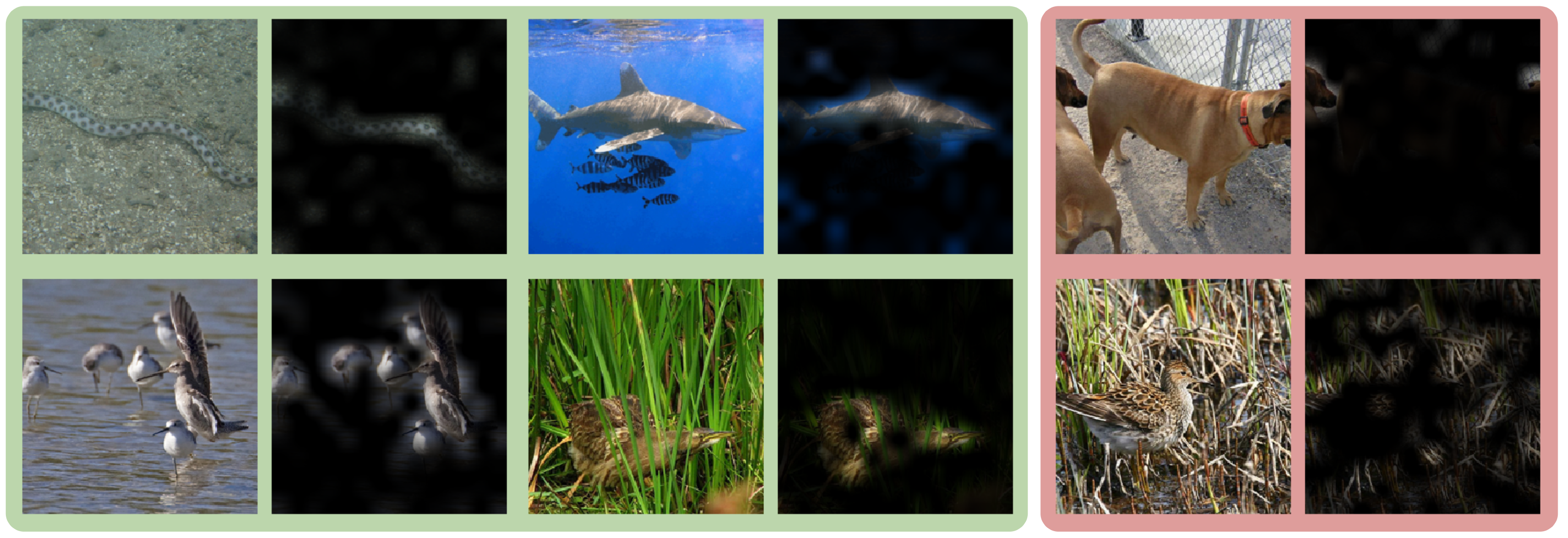}
    \caption{\textbf{Hydra Attention Visualization.} Visualization of the class token's Hydra attention in the last layer as specified in Sec.~\ref{subsec:hydra_vis}. The 4 images on the left are predicted correctly, while the two examples on the right are misclassified. In the top right image, the network focuses on the head of the wrong dog, guessing the wrong breed. Then on the bottom right, the network misses the bird completely. See the Appendix for more examples.}
	\label{fig:hydra_attn_vis}
\end{center}

\end{figure}

\subsection{Visualizing Hydra Attention} \label{subsec:hydra_vis}
One of the most desirable qualities of self-attention is its explainability: visualizing the focus of an attention-based model (e.g. with attention rollout \cite{attnrollout}) is typically straightforward. The same is less true for Hydra attention.

In order to visualize the focus of a Hydra attention module, we could construct attention matrices $\phi(Q)_h \phi(K)_h^T$ for $h \in \{1, \ldots, D\}$, but each would be rank 1 and it isn't clear how to combine $D$ different attention matrices when each is responsible for a different feature dimension. Simply averaging the heads together produces a meaningless result because each feature dimension encodes different information.

Instead, let's look at the information that each token contributes to the output for the class token. If we sample just the class token $c$'s output from Eq.~\ref{eq:hydra_attn}, we get
\begin{equation}
    \phi(Q)^c \odot \sum_{t=1}^T \phi(K)^t \odot V^t = \sum_{t=1}^T \phi(Q)^c \odot \phi(K)^t \odot V^t
\end{equation}
Thus, each token $t$ has a contribution to the output of the class token $c$ given by 
\begin{equation}
    \phi(Q)^c \odot \phi(K)^t \odot V^t
\end{equation}
To tell how this relates to the final prediction, we can use a method similar to Grad-CAM \cite{gradcam}: set the loss to be the logit for the predicted class, then obtain the gradient $g$ with respect to the output of the Hydra attention layer. Then the contribution of each token along the direction of that gradient is 
\begin{equation}
    (\phi(Q)^c \odot \phi(K)^t \odot V^t)^T g \label{eq:visualization}
\end{equation}
We plot this quantity for several different images in Fig.~\ref{fig:hydra_attn_vis} and the Appendix. For these visualizations, we normalize Eq.~\ref{eq:visualization} along the tokens and show the positive values. These focus maps show that while the math might be different, Hydra attention is performing much the same function as standard self-attention.

\begin{figure}[t]
\begin{center}
	\includegraphics[width=1\linewidth]{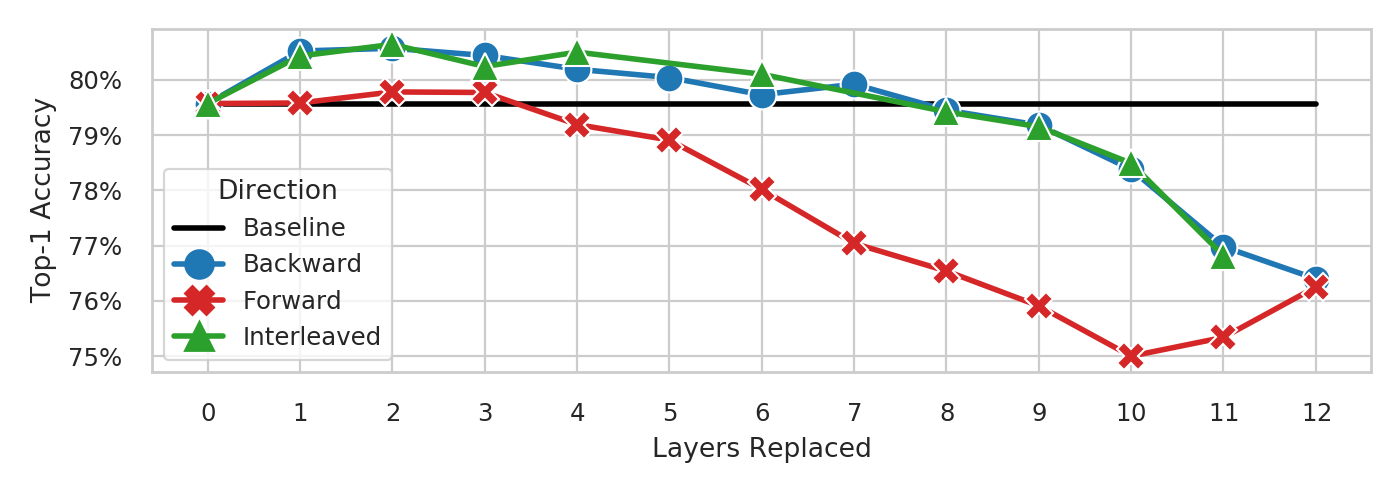}
\end{center}
\caption{\textbf{Which layers can we replace?} Replacing softmax self-attention with Hydra attention using different replacement strategies: from the front, from the back, or by interleaving the layers. In all cases, 0 indicates no layers replaced (the baseline), and 12 indicates that all layers were replaced. Surprisingly, with the right layer replacement strategy, Hydra attention can actually \textit{improve} accuracy on ImageNet by 1\%, while being being faster. Alternatively, we can replace up to 8 layers with no accuracy drop.}

\label{fig:backward_forward_scan}
\end{figure}
\subsection{Which layers can we replace?}
As discussed in Sec.~\ref{subsec:hydra_trick} and Sec.~\ref{subsec:kernel_choice}, Hydra Attention with a cosine similarity kernel mixes information between tokens in a different way to standard MSA \cite{attnisallyouneed}. Thus, it is perhaps unreasonable to replace every attention layer in the network with Hydra attention. In fact, Hydra attention creates a global feature from the tokens and applies that to each token weighted by $Q$. Because this is a global operation, it would make more sense in the later layers of the network, as at that point information has already been mixed locally. We test this in Fig.~\ref{fig:backward_forward_scan}, where we progressively replace the MSA attention layers in DeiT-B with Hydra attention following different strategies.

In this experiment, we observe that if we start replacing from the first layer of the network, the performance of the model quickly degrades. However, as it turns out, if we replace the layers in reverse starting with the last layer, we can actually \textit{improve} the accuracy of the model. And this improvement is so great that we can replace the last 8 layers of the network and still match the accuracy of the baseline DeiT-B model.

Then, if Hydra attention can be complementary with standard softmax attention, perhaps the best way to combine the two is to interleave them. In Fig.~\ref{fig:backward_forward_scan}, we also attempt to alternate MSA and Hydra layers following the principle that Hydra attention layers should follow MSA layers. However, we don't observe much tangible benefit to this interleaving strategy over starting at the back, suggesting that the number, not necessary the placement, of Hydra layers is what's important.

\begin{table}[t]
\newcommand{\bad}[1]{{\textcolor{red}{#1}}}
\newcommand{\good}[1]{{\textcolor{ForestGreen}{#1}}}
\begin{center}
	\begin{smalltable}{l c rr c rr c rr}
	    \toprule
	    Method  & $\qquad$ & \multicolumn{2}{c}{Accuracy (\%)} & $\qquad$ & \multicolumn{2}{c}{FLOPs (G)} & $\quad$ & \multicolumn{2}{c}{Speed (im/s)} \\
	    \midrule
	    Standard Attention \cite{attnisallyouneed} && 79.57 & && 17.58 & && 314.8 & \\
	    \midrule
	    AFT-Simple \cite{aft} && 74.02 & \bad{(-5.55)} && {\bf 16.87} & \good{(-4.0\%)} && 346.1 &  \good{(+9.9\%)} \\
	    PolyNL \cite{polynl} && 73.96 & \bad{(-5.61)} && {\bf 16.87} & \good{(-4.0\%)} && 353.8 & \good{(+12.4\%)}\\
	    \midrule
	    {\bf Hydra} (2 layers) && {\bf 80.64} & \good{(+1.1)} && 17.46 & \good{(-0.7\%)} && 321.9 & \good{(+2.3\%)} \\
	    {\bf Hydra} (8 layers) && 79.45 & (-0.12) && 17.11 & \good{(-2.7\%)} && 334.8 & \good{(+6.3\%)} \\
	    {\bf Hydra} (12 layers) && 76.40 & \bad{(-3.17)} && {\bf 16.87} & \good{(-4.0\%)} && 346.8 & \good{(+10.2\%)} \\
	    \bottomrule
	\end{smalltable}
    \caption{\textbf{Results.} Results for different attention methods in a DeiT-B \cite{deit} shell on ImageNet-1k \cite{imagenet} val trained on 224px images along with throughput measured on a V100. Hydra attention results in less accuracy drop than other $O(TD)$ attention methods (AFT-Simple \cite{aft} and PolyNL \cite{polynl}). Moreover, if we don't replace every attention layer in the network, Hydra attention can improve accuracy or keep it the same while still reducing FLOPs and increasing throughput. }
	\label{tab:results}
\end{center}

\end{table}

Note that other efficient attention methods such as AFT \cite{aft} and UFO-ViT \cite{ufovit} add conv layers instead of interspersing regular attention layers. Adding these convs serves much the same purpose as using self-attention to perform local mixing, but it's not clear whether the benefit of these prior methods come from the conv layers or their proposed attention layer. In this case, we've clearly isolated that Hydra attention can not only benefit the speed of the model, but also its performance. Future work may be interested in using convs instead.

\begin{table}[t]
\newcommand{\bad}[1]{{\textcolor{red}{#1}}}
\newcommand{\good}[1]{{\textcolor{ForestGreen}{#1}}}
\begin{center}
	\begin{smalltable}{l c rr c rr c rr}
	    \toprule
	    Method  & $\qquad$ & \multicolumn{2}{c}{Accuracy (\%)} & $\qquad$ & \multicolumn{2}{c}{FLOPs (G)} & $\quad$ & \multicolumn{2}{c}{Speed (im/s)} \\
	    \midrule
	    Standard Attention \cite{attnisallyouneed} && 81.33 & && 55.54 & && 92.5 & \\
	    \midrule
	    {\bf Hydra} (2 layers) && {\bf 81.92} & \good{(+0.59)} && 54.52 & \good{(-1.8\%)} && 96.3 & \good{(+4.1\%)} \\
	    {\bf Hydra} (7 layers) && 81.26 & (-0.07) && 51.96 & \good{(-6.4\%)} && 106.8 & \good{(+15.4\%)} \\
	    {\bf Hydra} (12 layers) && 77.85 & \bad{(-3.48)} && {\bf 49.40} & \good{(-11.0\%)} && 117.6 & \good{(+27.1\%)} \\
	    \bottomrule
	\end{smalltable}
    \caption{\textbf{384px Fine-Tuning.} Results for the models in Tab.~\ref{tab:results} fine-tuned with 384px images for 30 epochs. Even with more tokens, Hydra attention can still improve the accuracy over the baseline by 0.59\% with 2 layers and increase throughput by 15.4\% with 7 layers while matching the baseline's accuracy. }
	\label{tab:results384}
\end{center}

\end{table}
\subsection{Results} \label{subsec:results}
We present our final accuracy and FLOP count using Hydra attention in Tab.~\ref{tab:results} compared to standard $O(T^2D)$ attention and other $O(TD)$ methods on ImageNet-1k. Hydra attention achieves 2.4\% higher accuracy compared to other $O(TD)$ methods when replacing all layers. And when replacing fewer layers, Hydra attention can strictly outperform the baseline standard attention model: with 2 layers, accuracy increases by 1.1\% at 0.7\% reduced FLOPs and 2.3\% increase in throughput, and with 8 layers, accuracy stays the same with 2.7\% reduced FLOPs and 6.3\% faster throughput. Interestingly enough, the actual throughput increase outpaces the flops reduction substantially. This could be due to the observation in \cite{flashattn} that attention is memory-bound and because Hydra Attention uses less memory than standard attention.

\paragraph{Larger Images} To explore whether Hydra Attention retains these gains with more tokens, in Tab.~\ref{tab:results384} we fine-tune the backwards replacement models from Fig.~\ref{fig:backward_forward_scan} at a 384px resolution for 30 epochs using the hyperparameters suggested in \cite{deit}. This results in a model with almost 3 times the number of tokens, which should both accentuate the difference between $O(TD)$ and $O(T^2D)$ attention and indicate whether the global information propogation strategy of Hydra Attention is effective at these higher token counts. And indeed, in Tab.~\ref{tab:results384}, we see the same trend as with 224px images: Hydra Attention can increase accuracy by 0.59\% and throughput by 4.1\% with 2 layers or keep accuracy the same and increase throughput by 15.4\% with 7 layers this time.

\begin{table}[t]
\begin{center}
	\begin{smalltable}{cc c cr c cr c cr}
	    \toprule
	    \multirow{2}{*}{Image Size} &  \multirow{2}{*}{$\quad T \quad$} && \multicolumn{2}{c}{Baseline} && \multicolumn{2}{c}{Hydra} && \multicolumn{2}{c}{Local Window} \\
	    && $\qquad$& GFLOPs & Attn &$\qquad$& GFLOPs & Attn &$\qquad$& GFLOPs & Attn \\
	    \midrule 
	    224 & 197 && 17.6 & 4.10\% && 16.8 & 0.02\% && 17.6 & 4.10\%\\
	    384 & 577 && 55.1 & 11.13\% && 49.0 & 0.02\% && 51.1 & 4.10\%\\
	    448 & 785 && 78.0 & 14.56\% && 66.7 & 0.02\% && 69.5 & 4.10\%\\
	    1024 & 4097 && 657.3 & 47.06\% && 348.1 & 0.02\% && 362.8 & 4.10\%\\
	    1280 & 6401 && 1298.9 & 58.14\% && 543.8 & 0.02\% && 566.9 & 4.10\%\\
	    \bottomrule
	\end{smalltable}
    \caption{\textbf{FLOP Count vs Image Size.} FLOP count scaling of a ViT-B/16 model across different attention methods as image size increases. We also list the percent of total computation taken by creating and applying attention matrices. While Hydra attention significantly improves the FLOP count of the model at large image sizes, so does local window attention, which has already been shown effective on large images \cite{vitdet}. A limitation of Hydra attention is that it can only be 4\% faster than local window attention, though it's more general and can lead to proportionally higher throughputs.}
	\label{tab:flops}
\end{center}

\end{table}
\paragraph{Limitations}
Okay, but Hydra attention is 197x faster than standard attention (with $T=197$), so why is the maximum FLOP count reduction only 4\%? Well, it turns out that with ViT-B/16 $224 \times 224$ images ($T=197, D=768$), only 4.10\% of total model FLOPs reside in creating and applying attention matrices. With Hydra attention, this is reduced down to 0.02\%, essentially eliminating the cost of attention in the model. While this does result in a raw throughput increase of up to 10.2\% (see Tab.\ref{tab:results}), we can clearly do better.

Of course, the story changes as you increase the image size: in Tab.~\ref{tab:flops}, we repeat this computation for different image sizes, and the computation of standard attention balloons all the way up to 58\% with 1280px images, while Hydra attention remains negligible at 0.02\%. We test 384px images ourselves in Tab.~\ref{tab:results384}, and the speed-up for Hydra Attention is much more pronounced (up to a 27.1\% throughput increase). However, further work needs to be done to validate Hydra Attention on tasks that use more tokens (e.g. instance segmentation \cite{maskrcnn}). Though in those tasks, we'd be comparing against the local window attention used in ViTDet \cite{vitdet}, which has already been shown to be effective for large token regimes in images. Compared to local window attention, Hydra attention uses only 4\% fewer FLOPs at any image size, though its throughput would likely be proportionally higher (due to less memory usage).

In general, the usefulness of Hydra attention lies in its generality. Local window attention is a powerful solution for dense image prediction, but quickly becomes cumbersome with token sparsity (e.g., with masked pretraining \cite{mae,mae-video,videomae} or token pruning \cite{dynamicvit,spvit,adavit}). We leave this for future work to explore.

\section{Conclusion and Future Directions}
In this paper, we introduce Hydra Attention, an efficient attention module with many heads. We show that Hydra Attention outperforms other $O(TD)$ attention methods in Tab.~\ref{tab:kernel_choice} and can even work in tandem with traditional multihead self-attention to improve the accuracy of a baseline DeiT-B model in Fig.~\ref{fig:backward_forward_scan}. However, while Hydra attention works well on ImageNet classification (Tab.~\ref{tab:results}, Tab.~\ref{tab:results384}), its real potential for speed-up lies in larger images (Tab.~\ref{tab:flops}).

We've taken the first step in showing that Hydra attention can work at all and hope that future work can explore its use in other, more token-intensive domains such as detection, segmentation, or video. Moreover, Hydra attention is a general technique that doesn't make any assumptions about the relationships between tokens, so it can be applied to further improve the speed of token-sparse applications such as masked pretraining \cite{mae,mae-video,videomae} or token pruning \cite{dynamicvit,spvit,adavit}. We hope Hydra attention can be used as a step toward more powerful, efficient, and general transformers in the future.

\paragraph{Acknowledgements}
We want to thank Alexander Kirillov, Christoph Feichtenhofer, and Sean Bell for many thoughtful discussions and helpful comments!

\clearpage
%
%
\bibliographystyle{splncs04}
\bibliography{main}
\clearpage

\appendix

\begin{table}[b]
\begin{center}
	\begin{smalltable}{l c c r}
	    \toprule
	    Kernel & $\qquad\phi(Q)\qquad$ & $\qquad\phi(K)\qquad$ & Accuracy \\
	    \midrule
	    Cosine Similarity & \multicolumn{2}{c}{$x / ||x||_2$} & {\bf 76.37}\\
	    Tanh-L2 & $\text{tanh}(x)$ & {$x / ||x||_2$} & 76.17\\
	    Mean & \multicolumn{2}{c}{$x / \sqrt{T}$} & 75.95\\
	    CosSim + LN & \multicolumn{2}{c}{$x / ||x||_2$} & 75.74\\
	    Tanh-L2 + LN & $\text{tanh}(x)$ & {$x / ||x||_2$} & 75.22 \\
	    Tanh-Softmax & $\text{tanh}(x)$ & $\text{softmax}(x)$ & 74.18\\
	    Sigmoid-Softmax & $\sigma(x)$ & $\text{softmax}(x)$ & 74.02\\
	    L1 Normalization & \multicolumn{2}{c}{$x / ||x||_1$} & 70.75\\
	    \bottomrule
	\end{smalltable}
    \caption{\textbf{More Kernels.} We include other kernels we've tested here. Since most are asymmetric, we list $\phi$ for $Q$ and $K$ separately. }
	\label{tab:supp:kernel_choice}
\end{center}

\end{table}

\section{Other Kernels}
In Tab.~\ref{tab:supp:kernel_choice}, we list all the kernel functions (instantiations of $\phi(\cdot)$) we've tried with Hydra Attention. There are three main concerns we had while choosing these kernels. Namely, should $\phi(\cdot)$ be 1.) unbounded, 2.) allow for negative values, or 3.) be linear.

L2 normalization for cosine similarity (which is what we use in the paper), for instance, is bounded, allows for negatives, and is linear. Sigmoid, on the other hand, is bounded, is only positive, and is non-linear.

From our experiments, we've observed that while the function used for $Q$ is not that important, $K$ significantly benefits from being both linear and allowing negative values. Compared to L2 normalization, softmaxing $K$ significantly degrades performance. And out of the normalization techniques, we find L2 to work the best (over L1 or constant normalization).

Finally, since $\phi(K)$ (when not softmax) does not sum to 1, multiplying it by $V$ can produce magnitudes higher than standard attention. Thus, we thought it might be useful to normalize the result of the attention layer. We test two kernels with ``+ LN'', where the attention operation is followed by a Layer Norm (before the projection). This, however, does not seem to help, so it seems better to leave the kernel unnormalized here.

\section{More Visualizations}
\begin{figure}
    \centering
    \includegraphics[width=\linewidth]{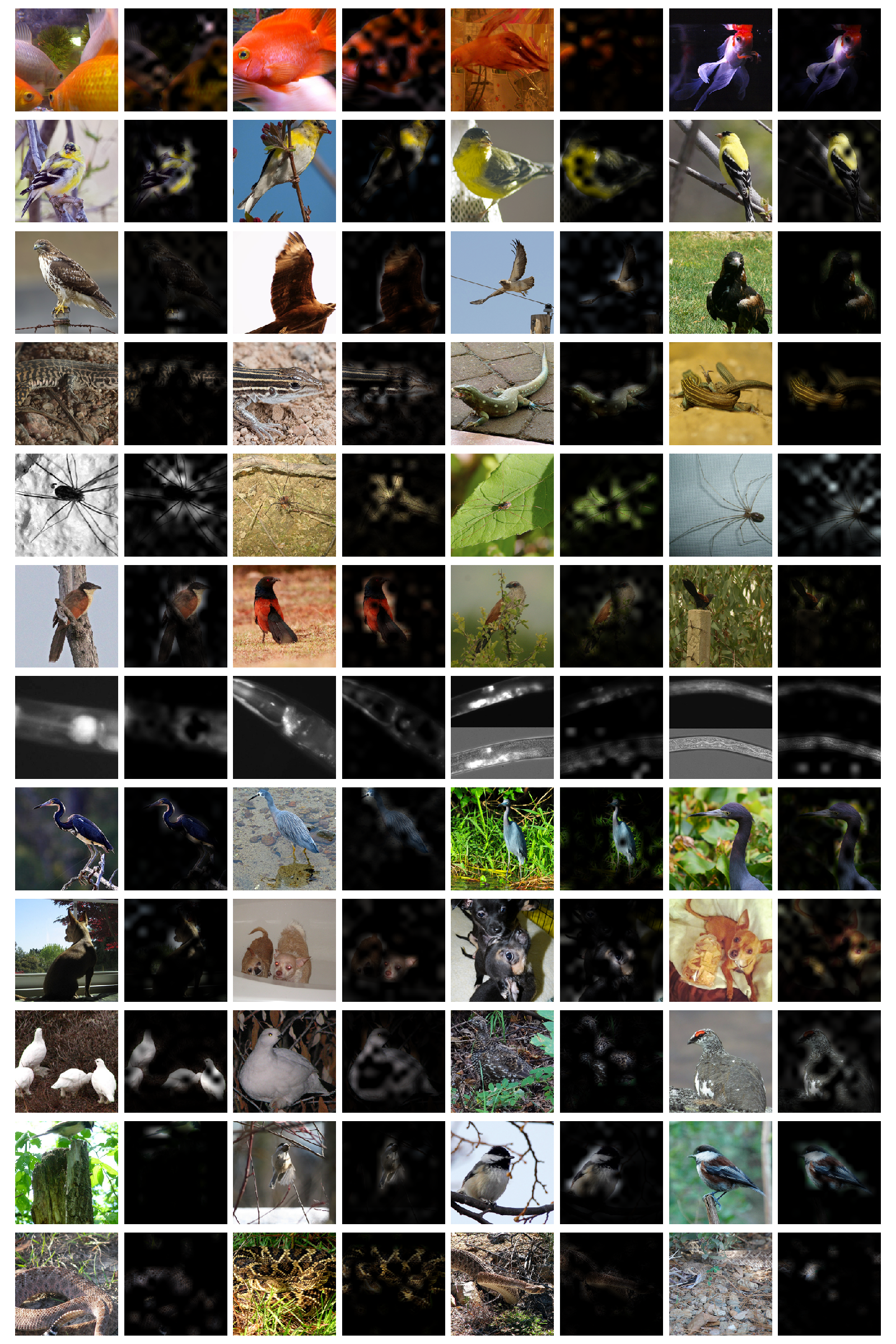}
    \caption{\textbf{More Visualization.} More visualization of the focus of the model using the method described in the paper.}
    \label{fig:supp:more_vis}
\end{figure}

In Fig.~\ref{fig:supp:more_vis}, we include several more image visualizations to supplement those in the main paper. These images were selected randomly from 12 different classes (4 per class) from the ImageNet-1k validation set with the only selection criteria being that the image is safe to view. The network predicts most of these images correctly.

\section{Code}
Hydra attention is extremely simple to implement. We give a reference implementation with the cosine similarity kernel here in PyTorch:

\begin{verbatim}
def hydra(q, k, v):
    """
    q, k, and v should all be tensors of shape
        [batch, tokens, features]
    """
    q = q / q.norm(dim=-1, keepdim=True)
    k = k / k.norm(dim=-1, keepdim=True)

    kv = (k * v).sum(dim=-2, keepdim=True)
    out = q * kv
    
    return out
\end{verbatim}

%
%

\end{document}